\documentclass[runningheads]{llncs}
\usepackage{graphicx}
\usepackage{xcolor}
\usepackage[misc,geometry]{ifsym}
\begin{document}
\title{Towards an IMU-based Pen Online Handwriting Recognizer}
\author{Mohamad Wehbi\inst{1}(\Letter) \and
Tim Hamann\inst{2} \and 
Jens Barth\inst{2} \and
Peter Kaempf\inst{2} \and
Dario Zanca\inst{1} \and
Bjoern Eskofier\inst{1}}

\authorrunning{M. Wehbi et al.}

\institute{Machine Learning and Data Analytics Lab \\
Friedrich-Alexander-Universität Erlangen-Nürnberg, Germany \\
\email{\{firstname.lastname\}@fau.de} \and
STABILO International GmbH, Germany\\
\email{\{firstname.lastname\}@stabilo.com}}

\maketitle              % typeset the header of the contribution
\begin{abstract}
Most online handwriting recognition systems require the use of specific writing surfaces to extract positional data. In this paper we present a online handwriting recognition system for word recognition which is based on inertial measurement units (IMUs) for digitizing text written on paper. This is obtained by means of a sensor-equipped pen that provides acceleration, angular velocity, and magnetic forces streamed via Bluetooth. 
Our model combines convolutional and bidirectional LSTM networks, and is trained with the Connectionist Temporal Classification loss that allows the interpretation of raw sensor data into words without the need of sequence segmentation. 
We use a dataset of words collected using multiple sensor-enhanced pens and evaluate our model on distinct test sets of seen and unseen words achieving a character error rate of 17.97\% and 17.08\%, respectively, without the use of a dictionary or language model.


\end{abstract}
\section{Introduction}

The field of handwriting recognition has been studied for decades, increasing in popularity with the advancements of technology. This increase in popularity is due to the substantial number of people using handheld digital devices that provide access to such technologies, and the desire of people to save and share digital copies of written documents. The aim of a handwriting recognition system is to allow users to write without constraints, then digitize what was written for a multitude of uses.

Handwriting recognition (HWR) is widely known to be separated into two distinct types, offline and online recognition \cite{plamondon2000online,priya2016online}. For offline recognition, a static scanned image of the written text is given as input to the system. Offline recognition, also known as optical character recognition (OCR), is the more common recognition technique used in a wide range of applications for reading specific details on documents, such as in healthcare and legal industry \cite{singh2012survey}, banking \cite{palacios2004handwritten}, and postal services \cite{srihari1993recognition}. 
Online handwriting recognition (OHWR), alternatively, requires input data in the form of time series and includes the use of an additional time dimension within the data to be digitized. 
This results in a dynamic spatio-temporal signal that characterizes the shape and speed of writing \cite{Kim2014}. OHWR systems are deployed in applications on tablets and mobile phones for users to digitize text using stylus pens or finger inputs on touch screens. Such systems use precise positions of the writing tip. However, one drawback that can be perceived is the need for a positional tracking system, whether it be a mobile touch screen, or any other pen tracking application. 
The need for such a tracking system restrains the user from writing on any surface and limits the usability of the system as well as the capability of the writer\cite{gerth2016handwriting}.

Another approach to applying OHWR is the use of inertial measurement units (IMUs) as, or integrated within, a writing tool. These tools provide movement data, such as accelerometer or gyroscope signals, which can be used for classification and recognition tasks. The major disadvantage of IMU sensors is that they are prone to error accumulation over time which, if not corrected, can lead to significant errors in the data recordings. Furthermore, IMU sensors generate noisy output signals, which is further intensified when touching a surface due to surface friction, 
which successively leads to lower performance at the task required due to deficient input data quality. 
However, when coupled with the correct models, IMUs produce beneficial data from which precise information can be extracted, such as the specific movements of a pen during handwriting. Moreover, IMUs are sourceless, self-contained and require no additional tools for data collection and extraction, and hence, a major advantage of IMU-based recognition systems is that no specific writing surface is required and systems rely only on the signals collected from the sensors.

In this paper, we discuss further the latter approach and introduce an OHWR system that uses sensor data recordings from pen movements to recognize writing on regular paper. We present an end-to-end system that processes sensor recordings in the form of time series data, and outputs the interpreted digital text on a tablet. Our system surpasses previous sensor-based pen systems in recognition rates, and is the first IMU-based pen recognizer that recognizes complete words and is not restricted to single character or digit recognition on paper. We use, as a digitizer, a regular ballpoint pen integrated with multiple sensors, and designed with a soft grip, that allows the user to write on a plain paper surface without constraints. 

The rest of the paper is structured as follows: Section \ref{section:RW} summarizes available OHWR systems, distinguishing between positional-based and IMU-based systems. Section \ref{section:DA} presents the digitizer used in our system and explains the data acquisition process. Section \ref{section:model} describes our end-to-end neural network architecture and describes the model training process, the hyperparameters used, and the data splitting. Section \ref{section:eval} reports the results and discusses the results obtained on distinct test sets. Section \ref{section:conc} outlines the future work to be implemented in our system.

\section{Related Work}\label{section:RW}

HWR has been a topic of interest in research for many years. Reviews about recognition systems  \cite{Kim2014,plamondon2000online} present pre-processing techniques, extracted features, in addition to different recognition models such as segment-and-decode methods and end-to-end recognition systems. In our work, we focus on the difference in the type of data used rather than methods of recognition. We briefly describe some previously developed recognition systems while distinguishing between systems using positional data and ones using IMU data.

\subsection{Positional-data based systems}

Basic OHWR systems were presented by \cite{bengio1995lerec,yaeger1998combining} in the late 1990s using Hidden Markov Model (HMMs) and Artificial Neural Networks (ANNs) that model the spatial structure of handwriting. A system developed using a multi-state time delay neural network was presented in \cite{jaeger2001online} using a dictionary of 5000 words with pen position and pen-up/pen-down data. Models in which both an image of the text along with pen tracking data were used to develop a Japanese handwriting recognition system \cite{jager2003state}.  Different language specific systems were implemented to apply recognition systems for different writing styles, such as Arabic \cite{tlemsani2018improved} and Chinese \cite{liu2004online}. 

The availability of public datasets considerably increased research in this field. The UNIPEN dataset \cite{guyon1994unipen} is a collection of characters with recorded pen trajectory information including coordinate data with pen-up/down features, which was used to implement a character recognition model using time delay \cite{unknown} and Convolutional \cite{mandal2019exploration} neural networks. An Arabic recognition system \cite{abd2014hmm} applied HMMs using the SUSTOLAH dataset \cite{musa2016towards}. 

The IAM-ONDB dataset \cite{liwicki2005iam} is considered the most popular dataset in the OHWR domain. It includes pen trajectories of sentences written on a smart whiteboard with an infrared device mounted on the corner of the board to track the position of writers’ pens, in addition to image data of written text from a collection of 86,272 word instances. \cite{liwicki2005iam} also introduced a HMM-based model with segmented data reaching up to about 66\% recognition rate. This rate increased to 74\% and consecutively to 79\% when recurrent neural networks (RNNs) and bi-directional Long Short-Term Memory-Networks (BLSTMs) were implemented with non-segmented data \cite{liwicki2007novel,graves2008novel}. An unconstrained recognition system was introduced in \cite{graves2008unconstrained} with the integration of External Grammar models, achieving a word error rate (WER) of 35.5\% using HMMs and 20.4\% using BLSTMs. The combination of diverse classifiers led to a word level accuracy of 86.16\% \cite{liwicki2011combining}.

A multi-language system \cite{keysers2016multi}, supporting up to 22 different scripts for touch enabled devices, was based on several components: character model, segmentation model, and feature weights. The model was trained and evaluated on different public and internal datasets, leading with error rates as low as 0.8\% to 5.1\% on different UNIPEN test sets, also achieving a character error rate (CER) and WER of 4.3\% and 10.4\%, respectively, on the IAM-ONDB dataset. The model results improved with the introduction of Bézier Curves, an end-to-end BLSTMs architecture, and language specific models with CER and WER of 2.5\% and 6.5\%, respectively \cite{carbune2020fast}.

The above metioned systems describe different methods for data pre-processing, feature extraction, and classification models, achieving different results for OHWR, while using input of either raw coordinate strokes or features extracted from these strokes. These systems were designed for position-based data that was extracted using specially designed hardware writing surfaces or touch screens and thus still pose a limitation if a system for digitzing paper-writing is required.

\subsection{IMU-data based systems}

The use of IMU data for HWR has been presented in different forms throughout the past years. 
Accelerometer-based digital pens for handwritten digit and gesture trajectory  were developed in \cite{wang2011accelerometer,jeen2013online} with an accuracy of 98\% \& 84.8\% over the ten digits, respectively. A 26 uppercase alphabet recognition system was developed using an inertial pen with a KNN classifier with 82\% accuracy~\cite{shaikh2015character}. Pentelligence \cite{schrapel2018pentelligence} combined the use of writing sounds with pen-tip motion from a digital pen equipped with microphones and IMU sensors for digit recognition reaching an accuracy of 98.33\% for a single writer. 

More recent studies used the Digipen \cite{koellner2019did} for the recognition of lowercase Latin alphabet characters. A recognition rate of 52\% was achieved using LSTMs. The Digipen \cite{wehbi2020digitizing} was also used for the classification of uppercase and lowercase Latin alphabet characters separately, with a different dataset than what was used in \cite{koellner2019did}. A 1-Dimensional Convolutional Neural Network (1D-CNN) model achieved an accuracy of 86.97\%. 

At the time of the development of these systems, no public dataset for this task was available for a concrete evaluation of different systems. The On-HW dataset \cite{ott2020onhw} was the first published IMU-based dataset and consisted of recordings of the complete Latin alphabet characters. It was released with baseline methods having an accuracy of 64.13\% for the classification of 52 classes. These results were based on the writer-independent scenarios described in the papers (when available), since a writer-dependent model is not a feasible model when developing an OHWR system for general use. 

For word level recognition, wearable technologies were implemented as approaches for HWR using IMUs. Airwriting is a tracked motion of continuous sensor stream, in which writing is of a single continuous stroke. It suffers from no surface friction and allows writing in free space. A digital glove equipped with accelerometers and gyroscopes for airwriting was designed in \cite{amma2012airwriting}, achiveing a WER of 11\% using an HMM model following the segment-and-decode approach, evaluated on nine users writing 366 words using a language model consisting of 60000 words. A CNN-RNN approach for in-air HWR \cite{gan2018unified} achieved a word recognition rate (WRR) 97.88\% using BLSTMs. Similar work was presented in \cite{gan2019air} with a recognition rate of 97.74\% using an encoder-decoder model. More recent work presented a wearable ring for on-surface HWR \cite{liu2020imu} which provided acceleration and the angular velocity data from the finger resulting in 1.05\% CER and 7.28\% WER on a dataset of 643 words collected by a single writer.

Differently from positional-based systems or wearable systems, IMU-based digital pen systems are still limited to single digit or character recognition, with a solution for word recognition not demonstrated yet. This is due to the fact that writing on a paper surface introduces a considerable amount of noise in the data which makes the learning of a recognition model challenging. Furthermore the evaluation of such systems has been conducted on very different setups and on limited data. Here we propose a system that aims at filling the gap between pen-based systems and other approaches.
We show that our IMU-based pen recognizer is practical for word recognition, and achieves significant improved results in comparison to previous pen devices.

\begin{figure}[t]
\centering
\includegraphics[width=\textwidth]{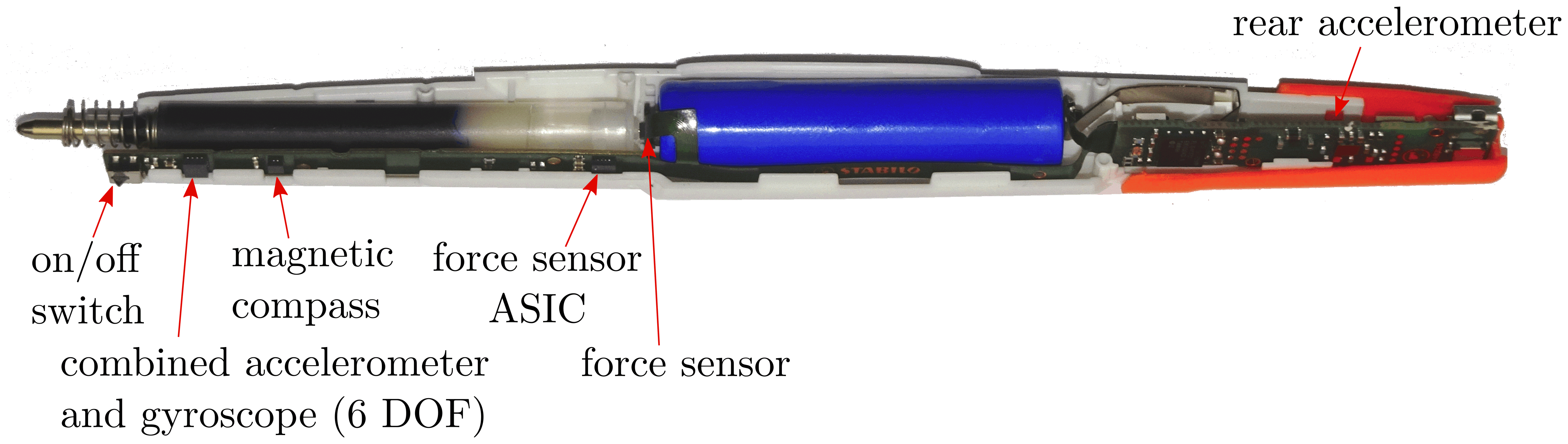}
\caption{The Digipen sensor placement} \label{fig2}
\end{figure}

\section{Data Acquisition \& Description}\label{section:DA}

In this section, we introduce the pen used as a digitzer and the data collection process, and describe the data that was used to train our model. We base our system on a set of digital pens of the same model to ensure that our work is not biased towards one single instance of the device. The pen model used in our system is the STABILO Digipen which was used in \cite{koellner2019did,wehbi2020digitizing,ott2020onhw}. 
The selection of this digitizer was based on the two main factors:

\renewcommand{\labelitemi}{$\bullet$}

\begin{itemize}

\item \textbf{Suitability}: 
The Digipen is a ballpoint pen that can be used to write on paper like any regular pen. It is equipped with five different sensors, a combined accelerometer and gyroscope module at one side of the pen close to the pen tip, another single accelerometer module close to the other end of the pen, a magnetometer module, in addition to a force sensor at the tip of the pen that provides data about when the tip touches a surface, displayed in Figure \ref{fig2}. This tool also includes a Bluetooth module that allows the transmission of collected sensor data to other devices in real-time. Accordingly, a trained recognition model can be integrated within a mobile app and can be used without the need for any further equipment but a mobile device. The Digipen streams sensor data via Bluetooth Low Energy with a sampling frequency of 100 Hz. Detailed information about the pen dimensions, sensor modules and ranges can be found in \cite{ott2020onhw,digisensors}.%Ergo, the pen fits well as a tool for our study when considering the aim of our work.

%\vspace{\baselineskip}

\item \textbf{Availability}:
The Digipen was not specifically designed for our work, and is available in a line of products. This implies that the system we developed is not for a single use case study, but can be further extended for different use case scenarios when required. Additionally, the availability of the pen allows the collection of data in a parallel manner which can accelerate a study that uses this tool. 

\end{itemize}

%\vspace{\baselineskip}

\subsection{Data Collection Application}

To collect the ground truth labels of the data, the Digipen is provided with a Devkit \cite{devkit} guide for the development of a mobile application for interaction with the pen. The application provides two files with similar timestamps of the recording session that can be used to extract data samples in the form of training data with the relative labels. A sample is defined as a complete word recording that consists of a series of timesteps.

\subsection{Data Recording}

Recording sessions were conducted in parallel using 16 different Digipens, with each session taking up to 45 minutes of recording time. A set of 500 words was used to collect the main set of data used for the system. Single words were displayed on the screen of a tablet, the users were asked to write the word on paper, in their own handwriting style, using the Digipen. 

The dataset included recordings from 61 participants who volunteered to contribute to our study, with some participants contributing less than the 500 required words due to time constraints. The number of samples collected was 27961 word samples. 

\begin{figure}[t]
\centering
\includegraphics[width=\textwidth]{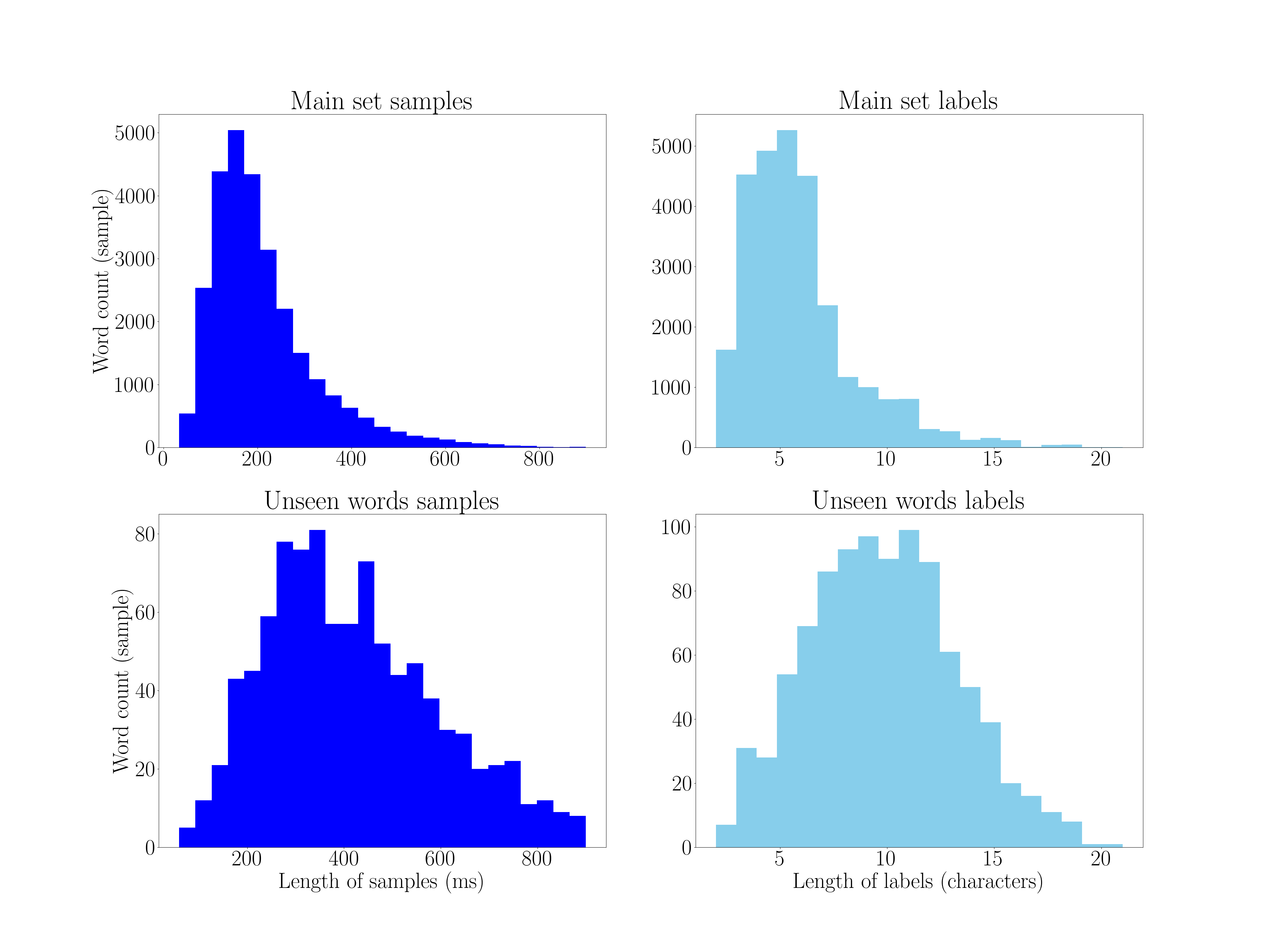}
\caption{A histogram displaying the number of samples in relation with (left) length of samples and (right) length of labels, in both (above) the main set and (below) the second test set.} \label{fig3}
\end{figure}

In addition to the main dataset, a separate dataset (unseen words set) was recorded from two other individuals. This recording consisted of random words selected from a set of 98463 words, different from the main set, serving as a second test set, with the purpose of testing the results of the system on unseen words. The final count of this set was 1006 sample recordings. Figure \ref{fig3} shows histograms of the data count of both datasets with respect to the lengths of the samples and labels separately.

\subsection{Data Preparation}

Data recording is subject to faults during the process. To ensure that our model was trained on valid data, all hovering data before and at the end of each single sample recording was trimmed out. This was achieved by removing the data associated with force sensor readings below a pre-specified threshold at the beginning and the end of a recording. From the first time this threshold was exceeded within that recording, all data was kept even when the force reading temporarily fell below that threshold. The threshold was determined experimentally through monitoring the highest force sensor values while hovering with the pen. Additionally, samples which appeared too short or long to be correct recordings were considered as faulty recordings and removed from the dataset. The data was then normalized per sample using the z-score normalization in order to input data features of a similar scale into the model. No further preprocessing or feature extraction was applied.

\section{End-to-end models}\label{section:model}

The architectures described in this section were inspired by research aimed at developing end-to-end recognition models in handwriting \cite{graves2008novel,carbune2020fast} and speech recognition \cite{feng2019end}. The use of Recurrent Neural Networks (RNNs), distinctively, Long Short-Term Memory networks (LSTMs), is common in the applications of handwriting and speech recognition due to the ability to transcribe data into sequences of characters or words while preserving sequential information. Bidirectional RNNs (BRNNs) make use of both past and future contextual information at every position of the input sequence in order to calculate the output sequences, and Bidirectional LSTMs (BLSTMs) have shown to achieve the best recognition results in the context of phoneme recognition when compared to other neural networks \cite{graves2005framewise}.

The models presented in this paper take input multivariate time series data samples of different lengths, comprised of 13 channels, representing the tri-axial measurements of the three IMU sensors and the magnetometer, in addition to the force sensor. 
In this section we evaluate different model architectures, describe the data splitting and the training process using raw sensor data.

\subsection{Model Architectures}

In the context of handwriting recognition using positional data, a model consisting of BLSTM layers proved sufficient to achieve the best recognition rates with the use of extracted feature vectors \cite{graves2008novel}, or resampled raw stroke data \cite{carbune2020fast}, while CNN models obtained the best character classification accuracy in systems using raw sensor data with the Digipen \cite{wehbi2020digitizing,ott2020onhw}. 

Following recent studies, we included the CNN model in our study. The model included four 1D-Convolutional layers, consisting of $1024, 512, 256, 128$ feature maps, consecutively, with kernel sizes of $5, 3, 3, 3$, respectively, and a fully connected layer of 100 units.

In contrast to positional data, in our case the input sequences are long due to a high sampling rate. Downsampling is not a viable option with IMU data because it leads to the loss of critical information \cite{bersch2014sensor}. Therefore, in addition to the CNN model, we implemented a CLDNN model (including Convolutional, LSTMs, and fully connected layers), which is typically used in speech recognition \cite{feng2019end}, where data samples are of high sampling rates and BLSTM models lead to latency constraints. 
The Convolutional layers reduce the dimensionality of the input features, which reduces the temporal variations within the LSTMs, which are then fed into the Dense layers where the features are tranformed into a space that makes that output easier to classify \cite{sainath2015convolutional}.
Hence, a CLDNN model allows to avoid latency constraints and slow training and prediction times which occur with BLSTM models. The model consisted of three Convolutional layers, followed by two BLSTM layers and a single fully connected layer. The Convolutional layers comprised of $512, 256, 128$ feature maps with kernel sizes of $5, 3, 3$, respectively. The BLSTM layers were of 64 units each, and the fully connected layer included 100 units. A grid search was implemented to determine the  optimal hyperparameters setup.

In both described models, Batch Normalization \cite{ioffe2015batch} and Max Pooling (of size 2) were applied after each Convolutional layer. The Relu activation was used in the Convolutional and the fully connected layers, while Tanh was used in the BLSTM layers. Random dropout \cite{srivastava2014dropout} with a dropout rate of 0.3 was applied after each layer to prevent overfitting and improve robustness of the system.

%A grid search was implemented to determine the best fitting hyperparameters for our models.
%Please notice that, a grid search was implemented to determine the  optimal hyperparameters setup for the model architectures described above. The number of Convolutional layers $n_c$ was considered between three and four layers for each model, i.e., $n_c \in \{3, 4\}$. The number of feature maps $n_f$ of the Convolutional layers ranged as $n_f \in \{1024, 512, 256\}$ in the first layer, then halving values per layer. Additionally, the kernel size ranged as $n_k \in \{7, 5, 3\}$  for the fist layer. The units of the BLSTM layers ranged $n_u \in \{128, 64, 32\}$ . Both models were tested with and without the fully connected layer, and were trained with increasing dropout values until no overfitting was perceived. To keep the models viable for real-time recognition, we used the hyperparameters that achieved the best complexity (prediction time) to performance ratio.

Similarly to the current developed systems in the field, we relied in our model on the Connectionist Temporal Classification (CTC) loss \cite{graves2006connectionist} with a Softmax output layer which provides an implicit segmentation of the data. The CTC is an RNN loss function that enables labeling whole sequences at once. It uses the network to provide direct mapping from an input sequence to an output label without the need of segmenting the data. It introduces a `blank' character that is used to find the best alignment of characters that best interprets the input.

\renewcommand{\arraystretch}{1.2} 
\begin{table}[t]
\centering
\caption{Number of samples in each set per fold (word samples).}\label{tab1}
\begin{tabular}{|p{2cm}|p{1cm}|p{1cm}|p{1cm}|p{1cm}|p{1cm}|}
\hline
& \multicolumn{5}{c|}{\textbf{Folds}} \\
\multicolumn{1}{|c|}{\textbf{Sets}} & \multicolumn{1}{c}{1} & \multicolumn{1}{c}{2} & \multicolumn{1}{c}{3} & \multicolumn{1}{c}{4} & \multicolumn{1}{c|}{5} \\
\hline
Training & 18452  & 18655  & 17588 & 17579 & 17598\\
Validation & 4614 &  4664  & 4397 & 4395 & 4400\\
Test & 4895 & 4642 & 5976 & 5987 & 5963\\
\hline
\end{tabular}
\end{table}

\subsection{Model Training}\label{section:train}

We split the main dataset into five folds, distributed into 49 users in the training set and 12 users in the test set, and train our model on the different folds separately. No writer appears in both sets to consider a writer-independent recognition task. The training data for each fold was divided into an 80/20 (training/validation) split. The unseen words dataset was used to test the effectiveness of the models for unseen word data. Table \ref{tab1} shows the different training, validation, and seen test sets, per each fold, not including any unseen words data.

For the implementation of our models, we used Keras/Tensorflow(v1) python libraries \cite{chollet2015keras,abadi2016tensorflow}, which include standard functions required for our work. The models were trained using a batch size of 64 samples and optimized with the Adam Optimizer \cite{kingma2014adam} with a starting learning rate of $10^{-2}$. A learning rate scheduler was implemented to monitor the validation loss and decrease the learning rate with a patience of 10 epochs and a factor of 0.8. We trained the models until the validation loss showed no decrease for 20 iterations after the minimum learning rate of $10^{-4}$ was reached, and saved the best model determined by the lowest validation loss during training. Finally, the evaluation of our model required the decoding of the CTC output into a word interpretation for which we used the Tensorflow standard CTC decoder function with a greedy search that returns the most likely output token sequence without the use of a dictionary.

\section{Evaluation \& Discussion}\label{section:eval}

Table \ref{tab3} presents the average results obtained. 
In terms of word recognition, the CNN model achieved the higher error rate of 35.9\% and 31.65\% average CER for seen and unseen words, respectively, which implies that the even though a CNN model achieved good results in character recognition \cite{ott2020onhw,wehbi2020digitizing}, it was not sufficient for the CTC to find the best character alignment within a word sample. The higher recognition rates were achieved by the CLDNN model, with an average of 17.97\% and 17.10\% CER. The models recognized unseen words without distinction from seen words, since the CTC learns to identify individual characters within the data. Additionally, having users in two distinct test sets different from users in the training sets provided a user-independent recognition model. 

Considering the different models in regard with the model complexity and time performance, Table \ref{tab3} shows the training time with respect to the trainable parameters of each model, in addition to the training iterations required to converge to the best performance. The CNN model consisted of a larger number of training parameters, however required lower training and prediction times. The CLDNN model achieved the better complexity to performance ratio with a significantly better recognition rate yet a longer training time relative to the CNN.

\begin{figure}[t]
\centering
\includegraphics[width=\textwidth]{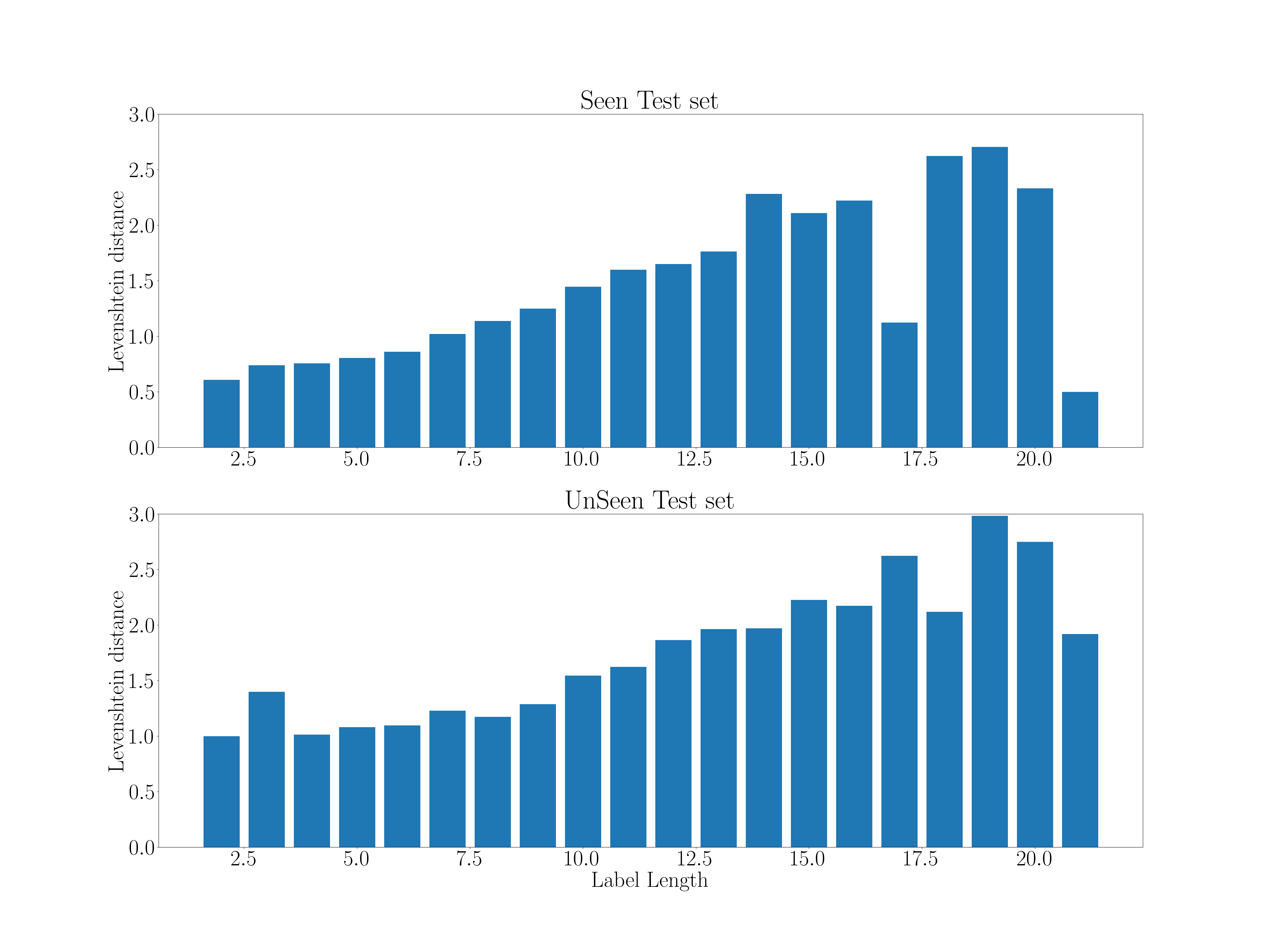}
\caption{A bar graph displaying the average Levenshtein Distance per label length for the seen and unseen test sets evaluated using the CLDNN model.} \label{fig4}
\end{figure}

In addition to the average CER, Figure \ref{fig4} reports the average Levenshtein Distance per label length for both test sets using the CLDNN model. This shows the minimum number of character edits, including insertions, deletions and substitutions, required to change a predicted word into the ground truth label. This means that the prediction of our model was on average divergent by $0.98$ and $1.66$ character edits for the average length $5.59$ and $9.72$ characters for the seen and unseen test sets, respectively. A detailed analysis of the errors showed that an average of 68\% of the predicted words were missing characters, which is due to cursive writing. 26\% of the prediction were of a substitution nature, which occurs between characters that look similar in both uppercase and lowercase, such as  `P-p', `K-k', and `S-s', while 6\% only included more characters than the relative ground truth, which occurs with multiple stroke characters.

\begin{table}[t]
\centering
\caption{Average error rates with respect to model properties. }\label{tab3}
\begin{tabular}{|l|c|c|c|c|c|}
\hline
\rule{0pt}{20pt} {\textbf{Models} }
& {\textbf{ \shortstack{Seen Words \\ Avg. \% CER}}} 
& {\textbf{ \shortstack{ UnSeen Words \\ Avg. \% CER }} } 
& { \textbf{ \shortstack{Avg. \\ Epochs} } } 
&  { \textbf{ \shortstack{Trainable \\ Parameters } } } 
& {  \textbf{ \shortstack{Seconds \\ per Epoch} } } \\

\hline
CNN     
&     35.90 ($\pm$ 2.01)    
&   31.65    ($\pm$ 1.07)  
&  153  
& 2,154,957
&   53     
\\
CLDNN  
& 17.97 ($\pm$ 1.98) 
&   17.10 ($\pm$ 1.68)
& 236   
& 743,373    
&     102 
 \\
\hline
\end{tabular}
\end{table}

The model used in our system followed the common used model in HWR systems, both offline and online, which is a stack of Convolutional or Recurrent layers trained with the CTC loss, and achieved an overall recognition rate similar to previous position-based models that did not make use of languages models \cite{graves2008novel}. However, this result is not directly comparable with previous systems, since these systems were trained on different data types, with sentence data, while our dataset consists of word data. Additionally, the state-of-the-art models in positional-based systems make use of complex language models. Moreover, the public IAM-OnDB dataset includes a higher number of classes in comparison to our dataset. Nonetheless, the presented results suggest that our system is on an adjacent level in terms of recognition rates without the use of a dictionary. 

Our system did not show the same level of recognition rates in comparison with the wearable systems described, which were trained on different datasets using distinct hardware. These systems followed the segment-and-decode approach with separate system-specific extracted features from uni-stroke data. Such systems provided air-writing capability, which does not fit for our paper-writing recognizer. The wearable ring presented in \cite{liu2020imu} was designed for on-surface writing, however, the system was developed and evaluated for a single specific writer. Also, writing with finger does not present the same efficiency in comparison with pen writing.

\begin{table}[t]
\centering
\caption{IMU-based pen recognizers with respective data type and performance, including the results of the presented CLDNN model. Digipen represents the pen used in this work. }\label{tab4}
\begin{tabular}{|c|c|c|c|}
\hline
\rule{0pt}{20pt} {\textbf{Model} }
& {\textbf{ \shortstack{ Data \\ Type }} } 
& { \textbf{ \shortstack{Recognition \\ Type} } } 
& {  \textbf{ \shortstack{Accuracy \\ (\%)} } } \\

\hline

\cite{jeen2013online}
&   Acc 
&   10 Digits
&  84.8
  
\\
\cite{wang2011accelerometer}
& Acc 
&  10 Digits
&  98

\\

\cite{schrapel2018pentelligence}
& Acc, Gyr, Sound
&   10 Digit
&    98.33

 \\
 
\hline 

\cite{koellner2019did}
& \textbf{Digipen}
&   26 Characters
&    52

\\

\cite{shaikh2015character}
& Acc, Gyr, Mag
&   26 Characters
&    82

\\

\cite{wehbi2020digitizing}
& \textbf{Digipen}
&   26 Characters
&    86.97 

\\
\hline

\cite{ott2020onhw}
& \textbf{Digipen}
&   52 Characters
&    64.13

\\
\hline
\textbf{Ours}
& \textbf{Digipen}
&  \textbf{Words}
&   \textbf{82.92 (CRR)} 

\\

\hline
\end{tabular}
\end{table}

Considering paper-writing recognition using sensor-equipped pens, our system achieved significant results in comparison to previously developed systems. Even though some previous systems used different hardware, our system, to the best knowledge of the authors, is the first IMU-based pen system that enables word recognition. Table \ref{tab4} shows a summary of the described sensor-equipped pens in Section \ref{section:RW}. Moreover, our model achieved an improved character recognition rate (CRR) by 18.79\% for the 52 Latin alphabet characters relatively to previous systems using the same hardware. %{\color{blue} The CRR is defined as the subtraction of the CER from the complete recognition rate.}

\section{Conclusion \& Future Work}\label{section:conc}

In this paper, we presented a system that applies OHWR by writing on normal paper using an IMU-enhanced ballpoint pen. We described the data collection tools and process in detail, and provided a complete system setup. We trained CNN and CLDNN end-to-end models that take normalized raw sensor data as input, and output word interpretations using the CTC loss with a greedy search decoder. The models were trained and evaluated using a five-fold cross-validation method, with test users being different from the users in the training set. We also evaluated the models on a separate test set to evaluate the efficiency of our system for unseen words. The presented CLDNN model showed the best performance without distinction between seen and unseen words. 

Our system showed significant improvements in comparison with previously presented character recognition systems using digital pens. With the results presented in this work, we showed that sensor-enhanced pens are efficient and yield promising results in the OHWR field in which digitizing writing on paper is required.
Accordingly, to further improve the applications of OHWR using digital pens, the dataset used in this work is planned to be published for use in the scientific community.
Future work following this will include complete sentence recognition, in addition to including digits and punctuation marks. 
Finally, the end-to-end model we presented requires minimal preprocessing, and mainly depends on the data, and thus to increase the robustness of a language recognizer, we plan to pair our model with a distinct dictionary or language model specific to the language to be recognized.

\subsubsection{Acknowledgments}
This work was supported by the Bayerisches Staatsministerium für Wirtschaft, Landesentwicklung und Energie as part of the EINNS project (Entwicklung Intelligenter Neuronaler Netze zur Schrifterkennung) (grant number IUK-1902-0005 // IUK606 / 002).
Bjoern Eskofier gratefully acknowledges the support of the German Research Foundation (DFG) within the framework of the Heisenberg professorship program (grant number ES 434/8-1).

\bibliography{refs.bib}
\bibliographystyle{splncs04}

\end{document}